\documentclass[sigconf]{acmart}
\usepackage{multirow}
\usepackage{xspace}
\usepackage{amsfonts, amsmath}
\usepackage[linesnumbered,ruled,vlined]{algorithm2e}

\usepackage{xcolor}
\definecolor{coolBlueGrey}{HTML}{4E79A7}
\definecolor{warmOrange}{HTML}{F28E2B}
\definecolor{softCoral}{HTML}{E15759}

\definecolor{mediumTeal}{HTML}{5DA5DA}
\definecolor{warmRed}{HTML}{F15854}
\definecolor{coolGreen}{HTML}{60BD68}

\definecolor{mediumLavender}{HTML}{A95AA1}
\definecolor{softYellow}{HTML}{DECF3F}
\definecolor{coolBlueGrey}{HTML}{4E79A7}

\newcommand{\paren}[1]{\left(#1\right)}
\newcommand{\set}[1]{\left\{#1\right\}}

\newcommand{\abs}[1]{\left|#1\right|}

\newcommand{\ds}{D_\textrm{seg}}
\newcommand{\dr}{D_\textrm{reg}}
\newcommand{\ls}{\mathcal{L}_\textrm{seg}}
\newcommand{\lr}{\mathcal{L}_\textrm{reg}}

\AtBeginDocument{
    
}

\copyrightyear{2023}
\acmYear{2023}
\setcopyright{rightsretained}
\acmConference[SC-W 2023]{Workshops of The International Conference on High Performance Computing, Network, Storage, and Analysis}{November 12--17, 2023}{Denver, CO, USA}
\acmBooktitle{Workshops of The International Conference on High Performance Computing, Network, Storage, and Analysis (SC-W 2023), November 12--17, 2023, Denver, CO, USA}
\acmDOI{10.1145/3624062.3625127}
\acmISBN{979-8-4007-0785-8/23/11}

\begin{document}

\title{Fast 2D Bicephalous Convolutional Autoencoder for Compressing 3D Time Projection Chamber Data}
\author{Yi Huang}
\email{yhuang2@bnl.gov}
\affiliation{%
    \institution{Brookhaven National Lab}
    \country{USA}
}

\author{Yihui Ren}
\email{yren@bnl.gov}
\affiliation{%
    \institution{Brookhaven National Lab}
    \country{USA}
}

\author{Shinjae Yoo}
\email{sjyoo@bnl.gov}
\affiliation{%
    \institution{Brookhaven National Lab}
    \country{USA}
}

\author{Jin Huang}
\email{jhuang@bnl.gov}
\affiliation{%
    \institution{Brookhaven National Lab}
    \country{USA}
}

\renewcommand{\shortauthors}{Huang et al.}

\begin{abstract}
High-energy large-scale particle colliders produce data at high speed in the
order of 1~terabytes per second in nuclear physics and petabytes per second
in high energy physics. Developing real-time data compression algorithms
to reduce such data at high throughput to fit permanent storage has drawn increasing attention. 
Specifically, at the newly constructed sPHENIX experiment at the Relativistic Heavy Ion Collider (RHIC),
a time projection chamber is used as the main tracking detector, 
which records particle trajectories in a volume of
three-dimensional (3D) cylinder. The resulting data are usually very sparse with
occupancy around $10.8\%$. Such sparsity presents a challenge to conventional
learning-free lossy compression algorithms, such as SZ, ZFP, and MGARD. The 3D
convolutional neural network (CNN)-based approach, \textit{Bicephalous
Convolutional Autoencoder} (BCAE), outperforms traditional methods both in
compression rate and reconstruction accuracy. BCAE can also utilize the
computation power of graphical processing units suitable for deployment in a
modern heterogeneous high-performance computing environment. This work introduces
two BCAE variants: BCAE++ and BCAE-2D. 
BCAE++ achieves a $15\%$ better compression ratio and a $77\%$ better
reconstruction accuracy measured in mean absolute error compared with BCAE. 
BCAE-2D treats the radial direction as the
channel dimension of an image, resulting in a $3\times$ speedup in compression throughput. 
In addition, we demonstrate an unbalanced
autoencoder with a larger decoder can improve reconstruction accuracy without
significantly sacrificing throughput. Lastly, we observe both the BCAE++ and BCAE-2D can benefit more from
using half-precision mode in throughput ($76-79\%$ increase) without loss in reconstruction accuracy.
The source code and links to data and pretrained models can be found at \url{https://github.com/BNL-DAQ-LDRD/NeuralCompression_v2}. 
\end{abstract}




\maketitle

\section{Introduction}
\label{sec:introduction}

High-energy particle accelerators, such as the Large Hadron Collider (LHC)~\cite{evans_lhc_2008} and the Relativistic Heavy Ion Collider (RHIC)~\cite{sPHENIX_TDR}, play a critical role in advancing knowledge about the fundamental building blocks of the universe. Particle accelerators work by accelerating charged particles close to the speed of light and colliding them, where they interact and produce new subatomic particles. Particle detectors are built around the collision point to detect these particle products, which encode the information about interactions at the collision. Basically, a tracking detector acts as a camera capturing three-dimensional (3D) particle trajectories. If there is a particle passing through it, each pixel or voxel will register an analog-to-digital (ADC) number above a zero-suppression threshold. This 3D ``camera'' can work with a large number of channels (millions to billions) at a high frame rate (10~kHz to GHz), producing a large amount of ADC data. 

\begin{figure}[ht]
    \centering
    \resizebox{.85\linewidth}{!}{\includegraphics{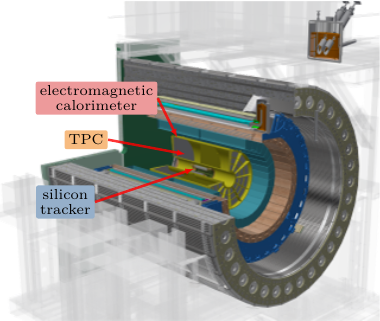}}
    \caption{Detector model.}
    \label{fig:tpc}
\end{figure}

Specifically, the recently constructed sPHENIX experiment at RHIC \cite{sPHENIX_TDR} consists of layers of tracking and calorimeter detectors aiming to study the microscopic nature of strongly interacting matter, ranging from nucleons to the strongly coupled quark-gluon plasma. Most sPHENIX data come from its main tracking detector, a Time Projection Chamber (TPC) (shown in {\autoref{fig:tpc}}). sPHENIX TPC digitizes 42M-voxels 3D pictures of the collision continuously at 77~kHz. Traditionally, to reduce and store the data in time, a filtering system, called \textit{level-1 trigger}, has been used. The trigger determines which data are more valuable, leading to a small subset of data being selected and stored for later analysis. 
Instead of a level-1 trigger, developing a high-throughput real-time compression algorithm to reduce and store \emph{all} collision signals has become increasingly important for future collider experiments with streaming data acquisition (DAQ) that aims to record all collisions~\cite{Bernauer:2022moy,AbdulKhalek:2021gbh}.

There is abundant literature in the lossy compression community, but few existing methods have been optimized or designed for sparse 3D TPC data. 
For example, the effectiveness of the error-bounded SZ~\cite{di_fast_2016,tao_significantly_2017,liu_optimizing_2022} compression algorithm method has been demonstrated in climate science and cosmology data. The fixed-rate compression method (ZFP)~\cite{lindstrom_fixed-rate_2014} has been motivated by hydrodynamics simulations, while the MultiGrid Adaptive Reduction of Data (MGARD)~\cite{ainsworth_multilevel_2019,liang_mgard_2022} method has been developed for compressing turbulent channel flow and climate simulation.
We hypothesize that most data challenges impacting the high-performance computing (HPC) community stem from distributed high-fidelity simulation in climate science, fluid dynamics, cosmology, and molecular dynamics. 
Therefore, by introducing this unique challenge from the particle accelerator community, 
we seek to ignite research interest in the scientific data reduction community.
Although all these compression algorithms have demonstrated reasonable performance with 3D TPC data, a specially designed neural network-based model, \textit{Bicepheoulous Convolutional Auto-Encoder} (BCAE)~\cite{huangTPCCompression}, can outperform them in both compression rate and reconstruction accuracy. 
However, as an initial proof-of-concept, BCAE has some drawbacks, including suboptimal compression throughput. 

This work proposes an improved version of BCAE, called \textit{BCAE++}, which improves the compression ratio from $27$ to $31$ and decreases the mean absolute error (MAE) from {$.198$ to $.112$}. MAE is an indicator of reconstruction accuracy---the lower, the better. 
We also introduce a two-dimensional (2D) variant of BCAE, \textit{BCAE-2D}, by replacing 3D CNN layers with a 2D one and treating the radial dimension of the TPC data as a channel dimension. 
This change has resulted in $3\times$ speedup in throughput. 
Because only the encoder portion of the auto-encoder neural network architecture will be used in real time 
and the decoder (decompression) can be used offline, 
this work also explores if expanding the number of decoder parameters can improve reconstruction accuracy. 
Lastly, we demonstrate a post-training ``trick'' using the half-precision representation of a network. 
It can enhance the throughput by over $70\%$ without losing reconstruction accuracy.
\section{TPC Data and BCAE Compression Methods}
\label{sec:method}

\subsection{TPC Data Preparation}
\label{subsec:data}

\autoref{fig:tpc} shows that the sPHENIX TPC is a perfect testbed for developing a high-throughput real-time compression algorithm. It is located between the inner silicon vertex tracker and the electromagnetic calorimeter.
Along the radial dimension, the TPC is composed of $48$ cylindrical layers of small sensors, which are grouped into three layer groups: inner, middle, and outer.
Each layer group has $16$ consecutive layers.
In the digitized data and for each TPC layer, the voxels are presented as a rectangular grid with rows along the $z$ (or horizontal) direction and columns along the azimuthal direction.
Within one layer group, all layers have the same number of rows and columns. 
This allows us to represent the ADC values from one layer group as a 3D array.
This study focuses on the outer layer group, where the array of ADC values has shape $(16, 2304, 498)$ in the radial, azimuthal, and horizontal orders. 

\begin{figure}[ht]
    \centering
    \includegraphics[trim=230 80 330 30, clip, width=.32\textwidth]{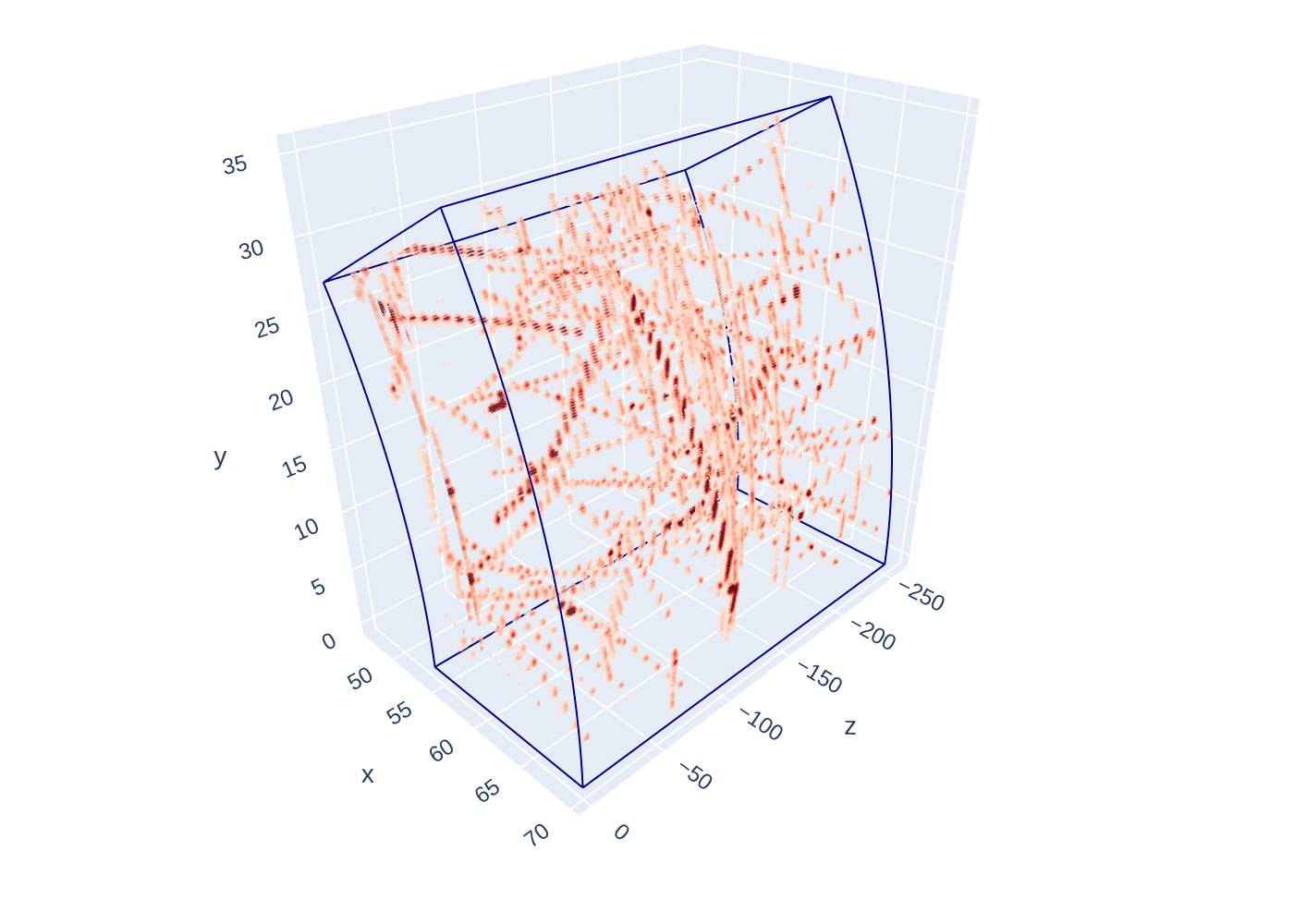}
    \caption{Example of a TPC Wedge. The $z$ axis is in the horizontal direction. The $x$ and $y$ axes are on the plane spanned by the radial and azimuthal directions.}
    \label{fig:tpcWedge}
\end{figure}

To match the subdivision of the TPC data assembly module in the readout chain, the voxel data are divided into $24$ equal-size non-overlapping sections:
$12$ along the azimuthal direction ($30$ degrees per section) and $2$ along the horizontal direction (divided by the transverse plane passing the collision point). 
We call one such section a \emph{TPC wedge} (\autoref{fig:tpcWedge}). 
The array of ADC values from each TPC wedge in the outer layer has shape $(16, 192, 249)$, listed in radial, azimuthal, and horizontal directions, respectively.
All ADC data from the same wedge will be transmitted to the same group of front-end electronics, 
after which a real-time lossy compression algorithm could be deployed.
Therefore, TPC wedges are used as the direct input to the deep neural network compression algorithms.

Here, we use the simulation data of $1310$ events for central $\sqrt{s_{NN}}=200$~GeV Au$+$Au collisions with 170~kHz pile-up collisions.
The data were generated with the HIJING event generator~\cite{Wang:1991hta} and Geant4 Monte Carlo detector simulation package~\cite{Allison:2016lfl} integrated with the sPHENIX software framework~\cite{sPHENIX_Software}. 
The simulated TPC readout (ADC values) from these events are represented in a $10$-bit unsigned integer $\in[0, 1023]$.
To reduce unnecessary data transmission between detector pixels and front-end electronics,
a zero-suppression algorithm has been applied. 
All ADC values below $64$ are suppressed to zero as most of them are noise.
This zero-compression makes the TPC data sparse at about $10\%$ occupancy (non-zero values).

We divide the $1310$ total events into $1048$ events for training and $262$ for testing.
Each event contains $24$ outer-layer wedges.
Thus, the training partition contains $25152$ TPC outer-layer wedges, while the testing portion has $6288$ wedges.
The compression algorithm aims to compress each wedge independently.

\begin{figure}[ht]
    \centering
    \includegraphics{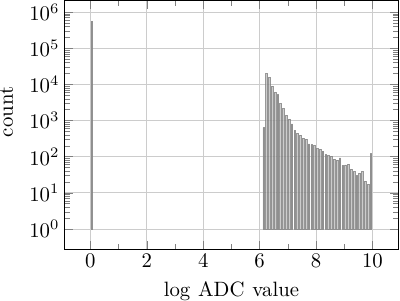}
    \caption{Log ADC distribution.}
    \label{fig:logADCDistr}
\end{figure}

Finally, as trajectory locations must be interpolated from neighboring sensors using the ADC values, it is important to preserve the relative ADC ratio between the sensors.
Hence, for this study, we trained autoencoders to reconstruct the \emph{log ADC values} ($\log_2(\text{ADC} + 1)$) instead of the raw ADC values.
A log ADC value is a float number in $[0., 10.]$.
Because of the zero-suppression at $64$, all nonzero log ADC values exceed $6$. 
The ground truth distribution of log ADC values is plotted in \autoref{fig:logADCDistr}.

\subsection{Bicephalous Convolutional Auto-Encoder (BCAE)}
\label{subsec:ae_with_two_decoder}

\begin{figure*}[ht]
    \centering
    \resizebox{.8\textwidth}{!}{\includegraphics{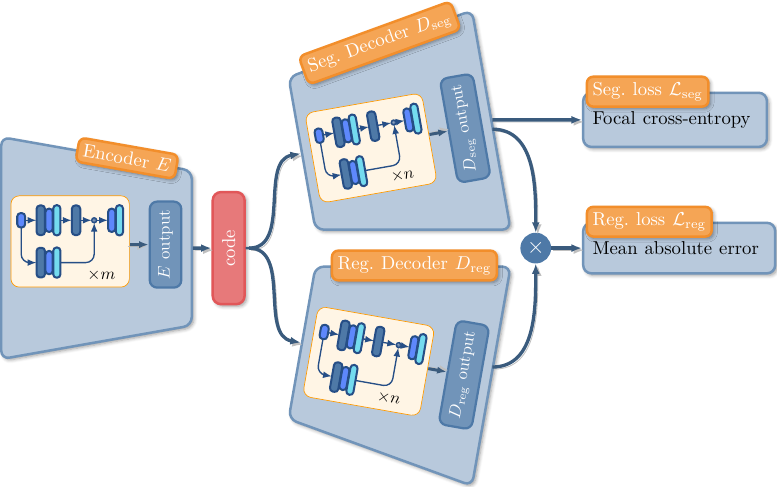}}
    \caption{\textbf{Bicephalous convolutional autoencoder (BCAE).}}
    \label{fig:BCAE}
\end{figure*}

An autoencoder~\cite{hinton_reducing_2006} is composed of one encoder and one decoder. 
The output from the encoder is called a \emph{code}.
Autoencoders are commonly used for data compression. The compression ratio is defined as the ratio between the size of an input and its code---the smaller the code, the higher the compression ratio.
The decoder takes in a code and produces a reconstruction of the input.
The distance between the original input and the reconstruction is used as the loss function to train the autoencoder network.
Although this approach may work well enough in cases where the input distribution is more regular (resembling a Gaussian distribution), it may struggle with a distribution such as those of zero-suppressed log ADC values~\cite{alanazi2020simulation,hashemi2019lhc}.

As evident in \autoref{fig:logADCDistr}, the log ADC value is bi-modal and has a sharp edge at $6.0$. 
For this irregular distribution, we need renovated autoencoder structures.
The BCAE~\cite{huangTPCCompression} was proposed as a potential solution.
As illustrated in \autoref{fig:BCAE}, in addition to the reconstruction decoder $\dr$, BCAE also has a segmentation decoder $\ds$ for voxel-wise bi-class classification. 
The segmentation decoder determines whether a voxel is zero (class $0$) or nonzero (class $1$).
The output produced by $\ds$ is assessed using the focal loss $\ls$, a specialized loss function designed to address imbalanced datasets~\cite{lin2017focal}. 
The output generated by the regression decoder $\dr$ is combined with that from $\ds$ to form the reconstruction of the input, which then is evaluated using a regression loss $\ls$. 
This study uses MAE for the regression loss.

Specifically, if we denote the total number of voxels by $M$ and let $\hat{l}_x$ be the output of $\ds$ for voxel $x$, the \textit{focal loss} is defined to be
\begin{multline}
    \ls\paren{\set{\left.\hat{l}_x\right|x};\gamma} = \frac{1}{M}\sum_{x}-l_x\log_2\paren{\hat{l}_x}\paren{1 - \hat{l}_x}^\gamma 
    \\ -\paren{1 - l_x}\log_2\paren{1 - \hat{l}_x}\paren{\hat{l}_x}^\gamma. \label{eq:clfLoss}
\end{multline}

Here, $l_x = 1$ if the voxel is positive and $0$ otherwise, and $\gamma$ is the focusing parameter.
We employ the focal loss because, on average, only $10.8\%$ of ADC values are nonzero.
In this study, we set the focusing parameter $\gamma$ to be $2$. 
Given a classification threshold $h$ and let $\hat{v}_{x}$ be the prediction for voxel $x$ produced by $\dr$,
the masked prediction $\tilde{v}_x$ is defined as $\tilde{v}_x = \hat{v}_x\mathbf{1}_{\hat{l}_x > h}$, where $\mathbf{1}$ is the characteristic function. 
Hence, the regression loss $\lr$ is defined as
\begin{equation}
    \lr\paren{\set{\left.\tilde{v}_x, v_x\right| x}; h} = \frac{1}{M}\sum_{x}\abs{\tilde{v}_x - v_x}. \label{eq:regLoss}
\end{equation}

Finally, to manage the gap between $0$ and $6$, we adopt another technique proposed by~\cite{huangTPCCompression} called \textit{regression output transformation}. We apply an output activation function $T(x) = 6. + 3. \exp(x)$ to the output from the regression decoder. Note that by applying $T$, \emph{all} regression output values are above $6.$, and the zero values in the reconstruction will result from the masking by segmentation output (refer to the definition of $\tilde{v}_x$).  

\subsection{BCAE++ and BCAE-HT}
\label{subsec:bcaepp}

Two modifications are made to the original BCAE~\cite{huangTPCCompression}. 
First, we pad the horizontal direction from length $249$ to length $256$ with zero. 
This makes halving the dimension more straightforward and enables using convolution/deconvolution with kernel size $4$, padding $1$, and stride $2$ uniformly throughout the encoder and decoder construction.
This change streamlines the neural network architecture search in a programmatic way.
In addition, this modification reduces the code dimension from $(8, 17, 13, 16)$ to $(8, 16, 12, 16)$ and improves the BCAE compression ratio from $27.041$ to $31.125$. 
Zero-padding in the horizontal direction is clipped during the evaluation, so reconstruction accuracy metrics are not inflated.
Second, we remove all the normalization layers in BCAE as they do not affect reconstruction performance significantly in a sufficiently long training. 
However, we can speed up training and inference without them. 
Based on these modifications, we constructed BCAE++ with a similar number of parameters to the original BCAE but with better performance (\autoref{tab:main}) and a larger compression ratio. 

We also introduce a high-throughput (HT) variation, \textit{BCAE-HT}.
The difference between BCAE++ and BCAE-HT is in the number of features (output channels) in the four residual blocks of their respective encoders. 
For BCAE++, the numbers of features are $8, 16, 32, 32$ (same as BCAE), while those for BCAE-HT are $2, 4, 4, 8$.   
The change reduces the encoder size of BCAE-HT to $5\%$ of those for BCAE and BCAE++ and yields a $76\%$ improvement in throughput. 

\begin{algorithm}[ht]
    \caption{\texttt{BCAE\_encoder\_2D}}
    \label{alg:enc_2D}
    \KwIn{number of blocks $m$, number of downsampling layers $d$}
    \KwOut{A \texttt{PyTorch} module}
    
    Initialize network $N$ to be an empty module list\;
    Append $N$ with $L_{\textbf{in}} = \texttt{Conv2D}\paren{i=16, o=32, k=7, p=3}$
    
    \For{$i \gets 1$ \KwTo $m$}{
        \If{$i \leq d$}{
            Append $N$ with $\texttt{AvgPool2D}\paren{k=2, s=2}$\;
        }
        Append $N$ with two residual block $\texttt{Res}\paren{i=32, o=32, k=3, p=1}$;
    }
    Append $N$ with $L_{\textbf{out}} = \texttt{Conv2D}\paren{i=32, o=16, k=1}$\;
    
    \Return{$N$}\;
\end{algorithm}

\subsection{BCAE-2D}
\label{subsec:2D}

Due to the thinness of the TPC wedge along the radial dimension---$16$ layers versus $192$ in the azimuthal and $249$ in the horizontal dimensions---it may be more reasonable to treat the layer dimension as channels of an ``image.''
Moreover, while the layers all have different radii, the number of columns (along the azimuthal direction) in one layer group remains the same. This means the distance between two adjacent voxels along the azimuthal direction is farther apart in an outer layer than an inner one.  
This breaks the inductive bias of translation invariance of a 3D convolution along the radial direction, making 2D convolution an even more appropriate choice for a TPC wedge. 

We detail the construction of the 2D encoder and decoder in \autoref{alg:enc_2D} and \autoref{alg:dec_2D}, respectively. 
The algorithms use $i$ and $o$ to denote the number of input and output channels, $k$ to show kernel size,
$p$ to indicate padding (default is $0$), and $s$ to signify stride (default is $1$).

\begin{algorithm}[ht]
    \caption{\texttt{BCAE\_decoder\_2D}}
    \label{alg:dec_2D}
    \KwIn{number of blocks $n$, 
          number of upsampling layers $d$, 
          output activation function $A$}
    \KwOut{A \texttt{PyTorch} module}

    \CommentSty{\# NOTE: a decoder must have the same number of upsampling steps as the downsampling steps in its corresponding encoder}
    
    Initialize network $N$ to be an empty module list\;
    
    \For{$i \gets 1$ \KwTo $n$}{
        \If{$i \leq d$}{
            Append $N$ with $\texttt{Upsample}\paren{\texttt{scale\_factor} = 2}$\;
        }
        Append $N$ with two residual block $\texttt{Res}\paren{i=32, o=32, k=3, p=1}$;
    }
    Append $N$ with $L_{\textbf{out}} = \texttt{Conv2D}\paren{i=32, o=16, k=1}$\;
    Append $N$ with output activation function $A$\;
    \Return{$N$}\;
\end{algorithm}

\begin{figure*}[ht]
    \centering
    \includegraphics{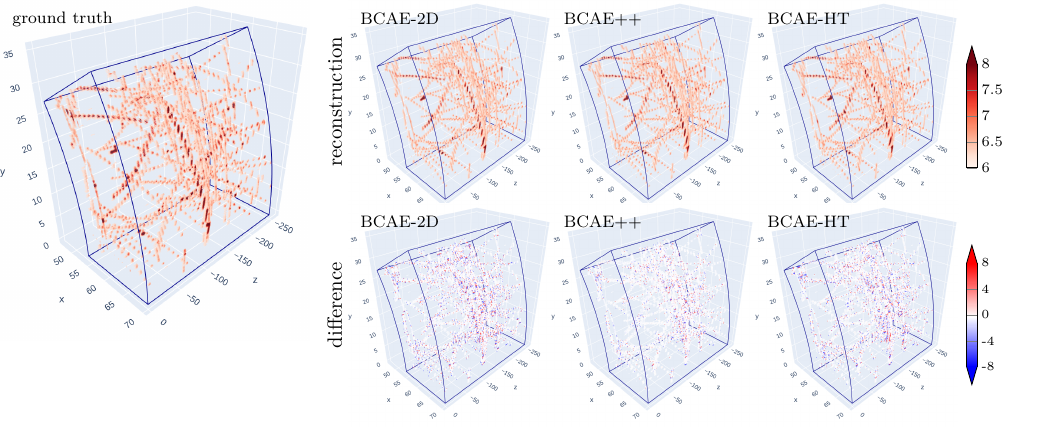}
    \caption{Reconstruction performance comparison.}
    \label{fig:reconstruction_notrans}
\end{figure*}

A BCAE-2D encoder $E$ is constructed using \texttt{BCAE\_encoder\_2D} with two parameters: the number of encoder blocks $m$ and number of downsampling layers $d$.
The segmentation decoder $\ds$ is constructed by \texttt{BCAE\_decoder\_2D} with \texttt{Sigmoid} as the output activation function $A$, while the regression decoder $\dr$ is constructed with the identity function as $A$. 
The number of upsampling layers $d$ in the decoders must be equal to the number of downsampling layers in their corresponding encoder. 
For simplicity in this study, we construct the two decoders in a BCAE-2D model with the same number of blocks ($n$). 
Hence, a BCAE-2D model can be denoted by BCAE-2D$(m, n, d)$, where $m$ is the number of encoder blocks in \autoref{alg:enc_2D}, $n$ is the number of decoder blocks in \autoref{alg:dec_2D}, and $d$ is the number of downsampling/upsampling layers. 

We keep $d$ at $3$ so the compression ratio of a BCAE-2D model is the same as the 3D variants. A BCAE-2D model with $d=3$ produces code with shape $(32, 24, 32)$. In Section~\ref{subsec:2dgrid}, we conduct a grid search on the numbers of encoder and decoder blocks. After balancing reconstruction accuracy and throughput, we choose BCAE-2D$(m=4, n=8, d=3)$ to represent the BCAE-2D models. Hence, in the sequel, we use BCAE-2D$(m=4, n=8, d=3)$ as the default BCAE-2D.

\subsection{Training Procedure}
\label{subsec:training}

We implement all BCAEs with \texttt{PyTorch 2.0}. 
The training is conducted on the $25152$ outer-layer TPC wedges in the training partition of the datasets, while $6288$ TPC wedges are reserved for testing.
We set the classification threshold $h$ in Equation~\eqref{eq:regLoss} to be $.5$ for both training and testing.
All BCAE models are trained with a batch size of $4$, and we train all BCAE++ and BCAE-HT for $1000$ epochs.
The initial learning rate is set at $10^{-3}$ and remains constant for the first $100$ epochs.
In the remaining epochs, we decrease the learning rate by $5\%$ every $20$ epochs. 
All BCAE-2D models are trained for $500$ epochs.
We set the initial learning rate at $10^{-3}$ and keep it constant for the first $50$ epochs.
In the remaining epochs, we decrease the learning rate by $5\%$ every $10$ epochs. 
For all BCAE models, we use the AdamW~\cite{loshchilov2017decoupled} optimizer with $\paren{\beta_1, \beta_2} = \paren{0.9, 0.999}$ and weight decay $0.01$. 

To improve the classification performance, we balance the contribution of the segmentation and regression loss dynamically as follows: assume the segmentation and regression losses at epoch $t$ are $\rho^{t}_\textrm{s}$ and $\rho^{t}_\textrm{r}$, respectively.
Denote the coefficient of the segmentation loss at epoch $t$ by $c_{t}$. 
Then, the coefficient for $\ls$ for epoch $t + 1$ is given by 
\[
c_{t + 1} = \frac{.5c_t + \paren{\rho^{t}_\textrm{r} / \rho^{t}_\textrm{s}}}{1.5}.
\] 
We set $c_0$ to be $2000$.

\begin{table*}[ht]
    \renewcommand{\arraystretch}{1.2}
    \centering
    \caption{Performance, encoder model size, and throughput comparison. The encoder size is measured in the number of trainable parameters. The reconstruction accuracy metrics and throughput are all measured with half-precision mode. The best performance with respect to each metric is underlined.}
    \begin{tabular}{l|r|r|r|r|r|r}
        \hline\hline
        model & MAE $\downarrow$ & PSNR $\uparrow$ & precision $\uparrow$ & recall $\uparrow$ & encoder size $\downarrow$ & throughput $\uparrow$ \\
        \hline
        BCAE-2D & $0.152$ & $11.726$ & $0.906$ & $0.907$ & $169.0$k & \underline{$\sim6.9$k} \\
        \hline
        BCAE++ & \underline{$0.112$} & \underline{$14.325$} & \underline{$0.934$} & \underline{$0.936$} & $226.2$k & $\sim2.6$k \\
        \hline
        BCAE-HT & $0.138$ & $12.376$ & $0.916$ &  $0.915$ & \underline{$9.8$k} & $\sim4.6$k \\
        \hline
        BCAE\cite{huangTPCCompression} & $0.198$ & $9.923$ & $0.878$ & $0.861$ & $201.7$k & $\sim2.4k$\\
        \hline\hline
    \end{tabular}
    \label{tab:main}
\end{table*}
\section{Result}
\label{sec:result}

\subsection{Compression Ratio}
\label{subsec:ratio}

The compression ratio is computed by the ratio between the input and the code.
Both input and code are treated as $16$-bit float.
As mentioned in Section~\ref{subsec:bcaepp} and \ref{subsec:2D}, the shape of the code produced by the BCAE-2D is $(32, 24, 32)$, and those for BCAE++ and BCAE-HT are both $(8, 16, 12, 16)$.
Because the TPC wedge has shape $(192, 249, 16)$, the compression ratio is $31.125$ for all newly introduced BCAE variants.
This is greater than the compression ratio of $27.041$ in the original BCAE~\cite{huangTPCCompression}.

\subsection{Comparing Encoder Model Size and Throughput}
\label{subsec:efficiencycomp}
The encoder model size is measured in the number of trainable parameters. 
Encoder throughput is measured in the number of TPC wedges processed per second. 
The input and output are allocated in GPU memory. Therefore, the file system input/output (IO) and host-to-device data transfer are not considered. 
All throughput experiments are conducted on a single \texttt{NVIDIA RTX A6000} GPU with driver version \texttt{535}. 
On the software side, we used \texttt{PyTorch 2.0} compiled with \texttt{CUDA 12.2}.

As shown in \autoref{tab:main}, BCAE++ has the largest encoder size of $226$k parameters, followed by the original BCAE with $202$k.
The BCAE-2D has $169$k but significantly higher throughput.
Reducing the number of BCAE++ parameters results in the BCAE-HT model size of $9.8$k.
Although better than BCAE++ ($2.6$k), the BCAE-HT's throughput ($4.6$k) is not as high as BCAE-2D ($6.9$k). 

\begin{figure*}[ht]
    \centering
    \includegraphics{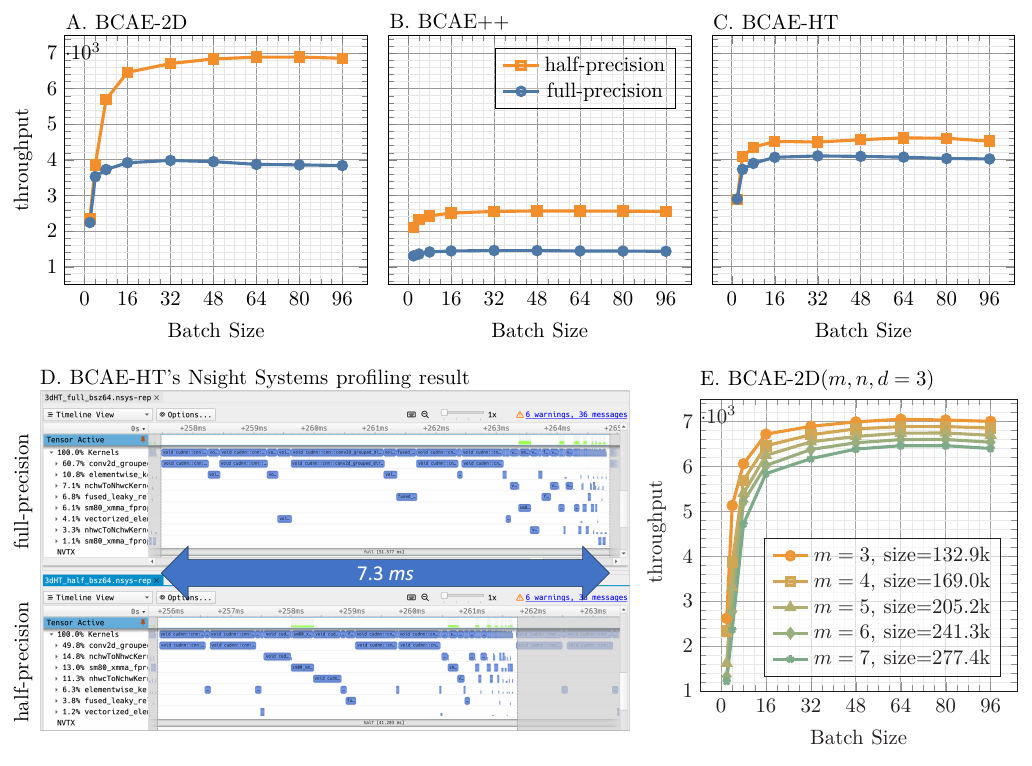}
    \caption{Panel A-C: Throughput in half- and full-precision modes on a single NVIDIA RTX A6000 GPU.
    Panel D: Diagnosing the lack of speedup by changing from full-precision to half-precision in the BCAE-HT model. This is due to small kernel sizes and the lack of \texttt{Tensor Core} activities. Panel E: Throughput in half-precision of BCAE-2D with $m=3, 4, 5, 6, 7$ encoder blocks and $3$ downsampling layers. The encoder size is measured in the number of parameters.}
    \label{fig:throughput} 
\end{figure*}

\subsection{Reconstruction Accuracy Comparison}
\label{subsec:perfcomp}
We evaluated performance of the BCAEs using four metrics: MAE, peak signal-to-noise ratio (PSNR), precision, and recall.  
Here, precision and recall are defined as follows:
\[
    \textbf{precision} = \frac{\sum_{x}\mathbf{1}_{\hat{l}_x > h} \mathbf{1}_{x > 6}}{\sum_{x}\mathbf{1}_{\hat{l}_x > h}}; \quad\textbf{recall} = \frac{\sum_{x}\mathbf{1}_{\hat{l}_x > h} \mathbf{1}_{x > 6}}{\sum_{x}\mathbf{1}_{x > 6}}.
\]

We compare BCAE-2D and BCAE++ performance in two computation modes shown in \autoref{tab:precision}. 
Given that compressing in half-precision yields negligible performance degradation while significantly boosting throughput, it is the most likely computation model for future deployment.
Hence, all BCAEs performance reported in \autoref{tab:main} are obtained with the half-precision computation mode. BCAE++ achieves the best scores in all reconstruction measurements.

\autoref{fig:reconstruction_notrans} compares the reconstruction performance of BCAE-2D, BCAE++, and BCAE-HT on one test TPC wedge. The noticeably different plots (second row) indicate the reconstruction produced by BCAE++ is the most accurate. 

\subsection{Investigating Half-precision Speedup}

Throughput is tested in two computation modes: full-precision and half-precision. 
In full-precision mode, the encoder weights and input are all set to $32$-bit floats.
In half-precision mode, we manually cast the encoder weights and input to $16$-bit floats.
In \autoref{fig:throughput}AB, for BCAE-2D and BCAE++, half-precision affords more than $70\%$ improvement in throughput.
In full-precision mode, BCAE-HT and BCAE-2D have a similar throughput of $4000$ frames per second.
However, BCAE-HT's speedup is much less (\autoref{fig:throughput}C).
This is due to the extremely small model size ($9.8$k) of BCAE-HT
after reducing the 3D convolution channel sizes 
from BCAE++'s $(8, 16, 32, 32)$ to $(2, 4, 4, 8)$. 
As shown in \autoref{fig:throughput}D, \texttt{Tensor Core} units are not used by those time-consuming convolution computations.

\begin{table}[ht]
    \renewcommand{\arraystretch}{1.2}
    \centering
    \caption{Reconstruction accuracy in full- and half-precision computation mode.}
    \label{tab:precision}
    \begin{tabular}{l|c|c|c|c}
        \hline\hline
        model & mode & MAE & precision & recall \\
        \hline
        \multirow{2}{*}{BCAE-2D}    & full & $0.151937$ & $0.905469$ & $0.906916$ \\
        \cline{2-5}     
                                    & half & $0.151965$ & $0.905326$ & $0.90705$ \\
        \midrule
        \multirow{2}{*}{BCAE++}     & full & $0.112347$	& $0.933817$ & $0.935779$ \\
        \cline{2-5}
                                    & half & $0.112342$ & $0.933852$ & $0.935741$ \\
        \midrule
        \multirow{2}{*}{BCAE-HT}    & full & $0.138443$ & $0.915891$ & $0.914562$ \\
        \cline{2-5}
                                    & half & $0.138441$ & $0.915780$ & $0.914701$ \\
        \hline\hline
    \end{tabular}
\end{table}

\begin{figure*}[ht]
    \centering
    \resizebox{\textwidth}{!}{\includegraphics{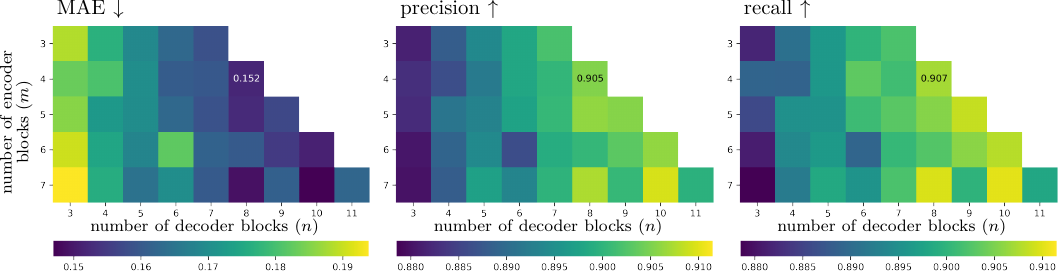}}
    \caption{Reconstruction accuracy of BCAE-2D models with varying encoder and decoder depths.}
    \label{fig:2d_grid}
\end{figure*}

\subsection{Investigating Auto-encoder Design}
\label{subsec:2dgrid}

Here, we study how the depth (the number of blocks) of a BCAE-2D model's encoder and decoders influence its reconstruction accuracy. For this purpose, we conduct a grid search on the number of encoder blocks ($m$ in \autoref{alg:enc_2D}), ranging from $3$ to $7$, and number of decoder blocks ($n$ in \autoref{alg:dec_2D}), from $3$ to $11$. \autoref{fig:2d_grid} illustrates the reconstruction accuracy of BCAE-2D models in MAE, precision, and recall. While the performance benefits significantly from deepening the decoders, the influence of encoder depth is relatively ambiguous. The benefit of a deep encoder is more obvious only when it is paired with decoders that are significantly deeper.
We also calculate the compression throughput in half-precision mode and demonstrate the result in \autoref{fig:throughput}E.
After balancing the reconstruction accuracy and compression throughput, we choose the BCAE-2D model with $m=4$ encoder blocks and $n=8$ decoder blocks to represent the BCAE-2D models.
\section{Conclusion and Discussion}
\label{sec:conclusion}

The TPC tracking detectors examined in this work represent a modern particle accelerator detector that produces a large volume of data at an extreme rate (1~TB/s to 1~PT/s). Unlike common simulation data from hydrodynamics, climate science, and cosmology, TPC data are zero-compressed and sparse, presenting a unique challenge to real-time data compression algorithms. We present two BCAE variants: BCAE++ and BCAE-2D. Compared to the BCAE, BCAE++ improves the compression rate by $15\%$ and reconstruction accuracy by $77\%$ measured in MAE. The BCAE-2D model treats the radial dimension of TPC data as a channel dimension and achieves a 3x speedup during inference compared to BCAE++. 

Based on this initial effort, there are several research directions worthy of future pursuit. For example, to further optimize the neural network throughput performance, we want to incorporate network pruning, quantization, and sparse CNN techniques. We also seek to extend our throughput comparison to include results of GPU-accelerated conventional lossy compression methods, such as MGARD-GPU and cuSZ~\cite{ tian_cusz_2020}. Finally, we anticipate this work can attract additional research interest in particle detector data compression by the scientific data reduction community.

\begin{acks}
This work was supported in part by Brookhaven National Laboratory under Laboratory Directed Research \& Development No.~23-048 and the Office of Nuclear Physics within the U.S.~DOE Office of Science under Contract No.~DESC0012704.
\end{acks}

\bibliographystyle{ACM-Reference-Format}
\bibliography{bib/bib}

\end{document}